# Computational Optimization, Modelling and Simulation: Recent Trends and Challenges

Xin-She Yang[a,*], Slawomir Koziel[b], and Leifur Leifsson[b]

[a]School of Science and Technology, Middlesex University, London NW4 4BT, United Kingdom
[b]Engineering Optimization and Modeling Center, School of Science and Engineering, Reykjavik University, 101 Reykjavik, Iceland.

**Abstract**

Modelling, simulation and optimization form an integrated part of modern design practice in engineering and industry. Tremendous progress has been observed for all three components over the last few decades. However, many challenging issues remain unresolved, and the current trends tend to use nature-inspired algorithms and surrogate-based techniques for modelling and optimization. This 4th workshop on Computational Optimization, Modelling and Simulation (COMS 2013) at ICCS 2013 will further summarize the latest developments of optimization and modelling and their applications in science, engineering and industry. In this review paper, we will analyse the recent trends in modelling and optimization, and their associated challenges. We will discuss important topics for further research, including parameter-tuning, large-scale problems, and the gaps between theory and applications.

## 1. Introduction

For any design and modelling purpose, the ultimate aim is to gain sufficient insight into the system of interest so as to provide more accurate predictions and better designs. Therefore, computational optimization, modelling and simulation forms an integrated part of the modern design practice in engineering and industry. As resources are limited, to minimize the cost and energy consumption, and to maximize the performance, profits and efficiency can be crucially important in all designs [1-6]. The stringent requirements of minimizing environmental impact and carbon footprint require a paradigm shift in scientific thinking and design practice. However, real-world problems are usually far more complex than models can capture and far more nonlinear than optimization tools can handle; consequently, approximations are necessity as well as a practical possibility.

Most design optimization typically involves uncertainty in material properties and parameters. In this case, optimal design does not necessarily mean robust. In fact, we often have to settle for the robust, suboptimal design options. After all, we wish to solve our design and modelling problems with sufficiently good accuracy assuming reasonable time expenditures [1,5].

Despite the significant progress made in the last few decades, many challenging issues still remain unresolved. Challenges may be related to various aspects and depend on many intertwined factors. In the current context, such challenges are related to nonlinearity, scale of the problem, time constraint and the complexity of the system. First, many problems are highly nonlinear, and thus their objective landscapes are multimodal. Consequently, multiple optima may be present. Many traditional algorithms do not cope well with such high multimodality. This necessitates new techniques to be developed. Second, many real-world problems may be very large-scale, though most optimization methods are tested over small-scale problems. Third, by far the most important factor concerning the solution process is the time constraint. Solutions have to be obtained within a reasonably time, ideally instantaneously in many applications, which poses additional challenges. Finally, the systems we try to model are usually very complex; however, we often use over-simplified models to approximate the true systems, which can introduce many unknown factors that affect the results and validation of the models [5,6].

The fourth workshop on Computational Optimization, Modelling and Simulations (COMS 2013) at the ICCS 2013 strives to provide an opportunity to foster discussion on the latest developments in optimization and modelling with a focus on applications in science, engineering and industry. In the remaining sections of this summary paper, we briefly review the recent trends, major challenges and discuss important topics for further research. We will also briefly introduce the topics and papers in this workshop.

---

\* Corresponding author, email: xy227@cam.ac.uk

## 2. Trends and Challenges in Computational Optimization

For proper formulation of optimization problems, the design objectives and behaviors of a system have to be re-formulated in mathematical terms to define an objective function (or functions) so that the formal relationship between the values of the designable parameters and the system performance can be established. In some cases, this relationship can be represented in a form of a scalar function that can be minimized, while in many other cases, a set of competing objectives can be only formulated, leading to a complex, multi-objective optimization problem. Even if the solution sets to a multi-objective problem can be found, it can result in a decision-making process to select the best combination out of a feasible set of, usually non-commensurable, objective sets. Such selection is not trivial, depending on the utility and/or decision criteria.

The recent trends in computational optimization move away from the traditional methods to contemporary nature-inspired metaheuristic algorithms [2,7,8], though traditional methods can still be an important part of the solution techniques. However, new studies and research tend to focus on the development of novel techniques that primarily based on swarm intelligence. New algorithms such as particle swarm optimization, cuckoo search and firefly algorithm have become hugely popular. One of the reasons for such popularity is that these metaheuristic algorithms are simple and easy to implement, and yet they can solve very diverse, often highly nonlinear problems. This partly meets the need to deal with nonlinearity in a non-conventional way.

Multimodality in many design problems mean the true global optimality is not easy to reach. In fact, there is no guarantee if the global optimality can be reached in a finite number of iterations. However, there is sufficient evidence that the global optimum can be found using nature-inspired algorithms in a vast majority of the cases. There are many reasons for good success rates in searching for optimality, a main reason is that metaheuristic algorithms use stochastic components or randomization techniques to increase the ergodicity of the iterative search path.

Nature-inspired algorithms have the advantages of simplicity, flexibility, and ergodicity [2,8]. These algorithms are typically very simple to understand and easy to implement, which requires little efforts for new users to learn. Therefore, researchers with diverse backgrounds can relatively easily use them in their own research. At the same, nature-inspired are quite flexible; that is, these seemingly simple algorithms can solve highly complex, high nonlinear optimization problems. In addition, nature-inspired metaheuristic algorithms can often find the global optimum solution within a relatively small, finite number of iterations. Some algorithms such as simulated annealing and cuckoo search can have guaranteed global convergence. That means that they can find the true global solution with a practically acceptable time scale. Such high ergodicity is common for the new nature-inspired algorithms, although it is not for the traditional algorithms such as gradient-based methods (unless in the special case of convex optimization where global optimality is also guaranteed).

To deal with the challenges of time constraint, the increase of computational efficiency and speed is crucially important. To reduce the solution time, a common technique is to use a low-fidelity model to approximate the true model, though strictly speaking there is no such thing as true models because all models are the approximations to the reality. However, for practical applications, most computationally extensive models can be approximated by computationally cheaper versions. An important issue is the accuracy that the approximate model can achieve. Typically, high-fidelity models tend to be computationally extensive, while low-fidelity models can speed up and thus reduce the overall computational costs [5,9-18]. However, there is always a trade-off between the accuracy of the approximate models and the computational costs.

There are a number of open problems regarding surrogate-based optimization. First, though surrogate-based methods have been successfully, they largely remain in the area of specialized engineering disciplines such as microwave engineering and part of aerospace engineering [18-20]. One of the reasons that may hinder the development in this area is that there is no specific guideline on how to construct the best surrogate-based models. The actual construction of an efficient surrogate still largely depends on the experience of the modelers/researchers and the specific knowledge of the subject. Often, most efficient models are physics-based, which requires even more specialized knowledge of the system. Further research can focus on the better approaches to achieve computationally cheap, high-accuracy models.

Despite rapid development and practical success of the new optimization methods, there are significant gaps between the theory and practice. In practice, many nature-inspired algorithm behave very well and they are usually efficient; however, there lacks theoretical understanding why these algorithm work well. In fact, apart from a few algorithms such as genetic algorithms with limited results on convergence, there is no mathematical proof for most algorithms why they converge and under what conditions. Therefore, there is a strong need for further research on the theoretical analysis of metaheuristic algorithms. This is also true for surrogate-based techniques, and the mathematical proof of convergence for certain specific methods and algorithms is yet to be seen.

Another significant gap is between large-scale and small-scale problems. In the current literature, most case studies and applications of nature-inspired algorithms are about small-scale to moderate scale problems with a dozen or at most a few hundred design variables. However, in real-world applications, problems are typically large-scale with thousands or even millions of design variables. It is not yet clear if the same methodology that works for small-scale problems can be extended to solve large-scale problems. One problem is the time constraint. Most problems cannot scale up linearly; consequently large-scale problems can be very computationally extensive. Further research can focus more efforts on large-scale, real-world problems.

## 3. Issues with Modelling and Simulation

Traditional modelling placed emphasis on mathematical modeling with most models based on partial differential equations. As the vast majority of mathematical models are not solvable analytically, approximate methods and numerical methods are the alternative. Unless the solution behaves smoothly, it may be intractable even with approximate methods. In this case, the only feasible approach is numerical solution. However, even though we can in principle solve a complex system numerically; this does not mean it is trivial in practice. In fact, most research efforts in the last few decades have dedicated to finding the most efficient methods in solving complex systems. As a result, numerical methods such as finite difference method, finite-element method and finite volume method have been developed [3]. They usually work very well for linear systems and weakly nonlinear systems.

However, for highly nonlinear, transient systems such as Navier-Stokes Equations, there is no truly efficient and accurate method. In many cases, linearization and approximations are often used, though the validity of such models is often questionable. In the context of computational optimization, models are often approximated by either using coarse-grid models or surrogate models.

There are many approximate methods such as perturbation methods, asymptotic methods, kriging, regression, trust-region reconstruction and surrogate-based methods. In recent years, the surrogate-based models have gained popularity, and one of the reasons is that it has flexibility to approximate fairly complex systems. These methods are particularly suitable for problems where the evaluation of the objective function is computationally expensive (e.g., 3D finite-element analysis of complex structures) and the optimization cost is a critical issue. In addition, a good combination is to use both surrogate-based simulation tool with an efficient metaheuristic optimization algorithm, and such hybridization could be even more powerful with suitable modifications using problem-specific knowledge.

There has been substantial research effort in the development of surrogate modelling methodologies that would allow to create models that are globally (or quasi-globally) accurate, smooth, and computationally cheap. While there are many different techniques available, both approximation-based (kriging [4], support-vector regression [8], neural networks [9]) and physics-based (space mapping [10]), many problems remain open, e.g., reducing the amount of data necessary to create the model, either by a smart sampling or by exploiting knowledge embedded in auxiliary, low-fidelity models. Typically, surrogate models are constructed using a sampling plan (i.e., design of experiments [4]) and nonlinear regression using, e.g., low-order polynomials [4] and kriging [13]. These approximation models are called functional surrogates and they are widely used in both academia and the industry [13]. For physics-based surrogates that exploit underlying low-fidelity models, numerous correction techniques are available, including simple response correction methods (both additive and multiplicative, see, e.g., [14]), space mapping (SM) [10], and shape-preserving response prediction (SPRP) [16]. Compared to surrogate-based optimization with functional surrogates, variable-fidelity optimization techniques can yield significant savings in computational cost, as the number of calls to the expensive simulation model is reduced, see, e.g., [17].

As mentioned earlier, the main challenge is to know how to construct the computationally efficient and yet sufficiently accurate models in a practical way with the ease for implementation, which still remains unresolved. As the speed of computers have increased steadily and the cost of a desktop steadily decreasing, current trends seem move to large-scale computation towards parallel computing, grid computing and cloud computing [21,22]. With vast computer resources to be harnessed for overnight computing, energy consumption may become another major issue. Computing has to go green, and in fact, green computing is a very active research area [23].

## 4. Recent Advances

Applications of optimization in engineering and industry are diverse and this is reflected in the papers submitted to this workshop. The responses and interests to our call for papers are overwhelming; however, due to limited space and time slots for presentations, many high-quality papers cannot be included in the workshop. The accepted papers of this workshop COMS 2013 at ICCS 2013 have spanned a wide range of applications and reflect a timely snapshot of the state-of-the-art developments in computational optimization, modeling and simulation.

For the algorithm developments, Yang et al. introduce a multiobjective flower algorithm for optimization, while Thaher et al. present a study of the maximum convex sum algorithm with the application in determining environmental variables. For applications and studies in optimization, Koziel et al. present a detailed study of shape-preserving response prediction for design optimization, while Zelazny et al. study the bicriteria flow optimization scheduling problems using simulated annealing. Koziel and Leifsson optimize airfoil shape using multi-level CFD-based low-fidelity model selection. In addition, Blum et al. solve the 2D bin packing problems by means of a hybrid evolutionary algorithm. Xavier et al. use genetic algorithm for solving history matching problems, and Trunfio et al. carry out GPU-accelerated optimization for mitigating wildfire hazard.

On the modelling and simulation developments, Koziel et al. use physics-based surrogates for low-cost modeling of microwave structures, while Zimarez use a genetic algorithm to model vasculature of a dicotyledon leaf. Conde et al. model sketch arm and custom closets for rapid prototyping systems. Furthermore, Sawi et al. use a small-world network model for simulating targeted attacks, while Baluja uses neighborhood preserving codes for applications in stochastic search.

From the above studies, we can see that the applications of computational optimization and modelling are diverse and wide-ranging. There is no doubt that more and more applications will appear in the near future. This workshop provides a timely platform for further discussions and development in optimization and modeling.

## 5. Conclusions and Open Questions

There are tremendous progress and activities in computational optimization, modelling and simulation. New trends start to shape the research landscape in the above areas. Current trends with more active research can be summarized as the following areas:
- Nature-inspired metaheuristic algorithms,
- Surrogate-based model and optimization,
- Large-scale problems,
- Green computing and grid computing.

However, there are many important issues that still motivate researchers to search for better algorithms and efficient surrogate techniques. For example, the performance of an algorithm may closely depend on the parameter settings of its algorithm-dependent parameters. Some attempts in the literature is to either use chaos to increase the diversity and ergodicity of an algorithm [24] or to use parameter-tuning methods to influence the convergence of an algorithm [25], which leads to active research activities. We can summarize the challenges in modelling and optimization as the following open questions:
- What exactly control the performance of a metaheuristic algorithm and its convergence rate?
- How to make an algorithm truly intelligent?
- How to optimally balance the local search and global search capabilities in an algorithm?
- Will the methodology for small-scale problems scale up and works equally well for large-scale problems?
- What is the best way to construct a good surrogate model for a given problem?

- What is the best choice of algorithms and surrogate models for a given problem?

These important questions can form an important set of active research topics for the next few years. Any insight gained may significantly alternate the research path and landscape in modelling and optimization.